%% file: acl2020.tex
\documentclass[11pt,a4paper]{article}
\usepackage[hyperref]{acl2020}
\usepackage{subcaption}
\usepackage{latexsym}
\usepackage{graphicx}
\usepackage{times}
\usepackage{amsmath}
\usepackage{amsfonts}
\usepackage{latexsym}
\usepackage{multirow}
\usepackage{paralist}

\usepackage{microtype}

\aclfinalcopy 
\def\aclpaperid{1050} 

\title{Mitigating Gender Bias Amplification in Distribution by \\  Posterior Regularization}

\author{Shengyu Jia$^{\clubsuit}$\thanks{\ \ Both authors contributed equally to this work and are listed in alphabetical order.}, Tao Meng$^{\spadesuit}$\footnotemark[1], Jieyu Zhao$^\spadesuit$, Kai-Wei Chang$^\spadesuit$ \\
$^\clubsuit$ Tsinghua University \\
$^\spadesuit$ University of California, Los Angeles  \\
  {\tt jiasy16@mails.tsinghua.edu.cn, } \\
  {\tt \{mengt18, jieyuzhao, kwchang\}@ucla.edu} \\
}

\date{}

\begin{document}
\maketitle

\begin{abstract}
    Advanced machine learning techniques have boosted the performance of natural language processing. Nevertheless, recent studies, e.g., \citet{jieyu2017men} show that these techniques inadvertently capture the societal bias hidden in the corpus and further amplify it.
    However, their analysis is conducted only on models' top predictions. In this paper, we 
    investigate the gender bias amplification issue from the distribution perspective and demonstrate that the bias is amplified in the view of 
    predicted probability distribution over labels. We further propose a bias mitigation approach based on posterior regularization. With little performance loss, our method can almost remove the bias amplification in the distribution.
    Our study sheds the light on understanding the bias amplification. 
\end{abstract}

\section{Introduction}    
    %
    Data-driven machine learning models have achieved high performance in various applications. Despite the impressive results, recent studies  (e.g., \citet{wang2019balanced,hendricks2018women}) demonstrate that these models may carry societal biases exhibited in the dataset they trained on. 
    In particular,  \citet{jieyu2017men} show that a model trained on a biased dataset may amplify the bias.
    For example, we can consider a task of labeling the activity and objects depicted in an image. The training set contains 30\% more images with ``woman cooking'' than ``man cooking''. However, when evaluating the top predictions of a trained model, the disparity between males and females is amplified to around 70\%. 
    Based on this observation, \citet{jieyu2017men} conduct a systematic study and propose to calibrate the top predictions of a learned model by injecting corpus-level constraints to ensure that the gender disparity is not amplified. 
    
    
    
    However, when analyzing the top predictions, the models are forced to make one decision. Therefore, even if the model assigns high scores to both labels of ``woman cooking'' and ``man cooking'', it has to pick one as the prediction. This process obviously has a risk to amplify the bias. However, to our surprise, we observe that gender bias is also amplified when analyzing the posterior distribution of the predictions. Since the model is trained with regularized maximal likelihood objective, the bias in distribution is a more fundamental perspective of analyzing the bias amplification issue. 
    
    
    In this paper, we conduct a systematic study to quantify the bias in the predicted distribution over labels. Our analysis demonstrates that when evaluating the distribution, though not as significant as when evaluating top predictions, the bias amplification exists. About half of activities show significant bias amplification in the posterior distribution, and on average, they amplify the bias by 3.2\%.  
    
    We further propose a new bias mitigation technique based on posterior regularization because the approaches described in \citet{jieyu2017men} can not be straightforwardly extended to calibrate bias amplification in distribution. 
    With the proposed technique, we successfully remove the bias amplification in the posterior distribution while maintain the performance of the model. Besides, the bias amplification in the top predictions based on the calibrated distribution is also mitigated by around 30\%. These results suggest that the bias amplification in top predictions comes from both the requirement of making hard predictions and the bias amplification in the posterior distribution of the model predictions. 
    Our study advances the understanding of the bias amplification issue in natural language processing models. The code and data are available at \url{https://github.com/uclanlp/reducingbias}. 
    

    
\section{Related Work}
    \paragraph{Algorithmic Bias}
    Machine learning models are becoming more and more prevalent in the real world, and algorithmic bias will have a great societal impact \citep{tonry2010social,buolamwini2018gender}. Researchers have found societal bias in different applications such as coreference resolution \citep{rudinger2018gender,zhao2018gender}, machine translation~\cite{stanovsky2019evaluating} and online advertisement \cite{sweeney2013discrimination}. Without appropriate adjustments, 
    the model can amplify the bias~\cite{jieyu2017men}. Different from the previous work, we aim at understanding the bias amplification from the posterior perspective instead of directly looking at the top predictions of the model.
    
    \paragraph{Posterior Regularization} 
    The posterior regularization framework~\cite{ganchev2010posterior} is aiming to represent and enforce constraints on the posterior distribution. It has been shown effective to inject domain knowledge for NLP applications. For example, \citet{ji2012question, gao2014learning} design constraints based on similarity to improve  question answering and machine translation, respectively.
    \citet{yang2014context} propose constraints based on lexical patterns in sentiment analysis.
    \citet{meng2019target} apply corpus-level constraints to guide a dependency parser in the cross-lingual transfer setting. In this paper we leverage corpus-level constraints to calibrate the output distribution. Our study resembles to the confidence calibration \citep{guo2017calibration,pakdaman2015obtaining}. However, the temperature turning and binning methods proposed in these papers cannot straightforwardly be extended to calibrate the bias amplification. 
    

\section{Background}
\label{sec:background}
\begin{figure}[t]
        \centering
        \includegraphics[width=\linewidth]{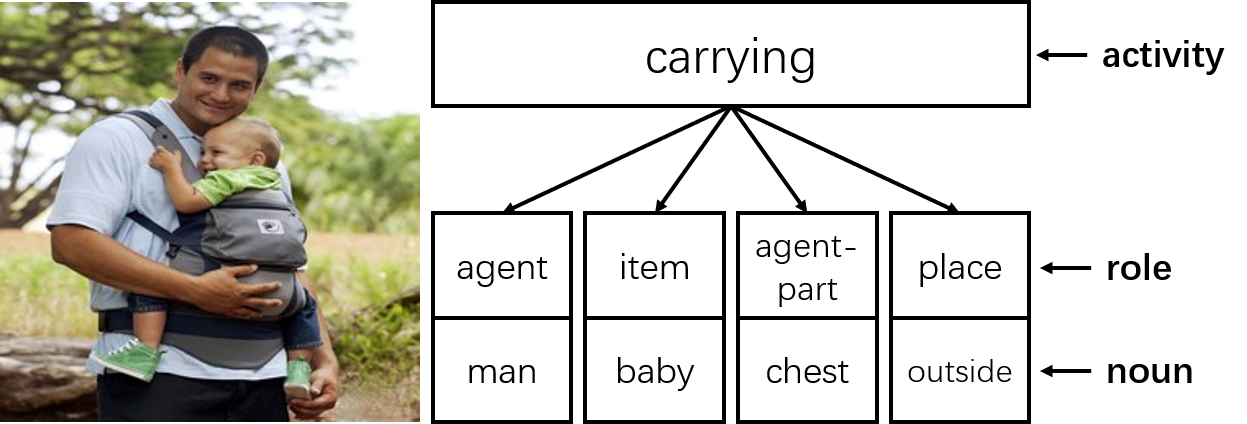}
        \caption{An instance from the imSitu dataset. Given an input image, the task it to identify the activity depicted in the image as well as the objects (noun) and their semantic role.}
        \label{fig:imsitu}
    \end{figure}

    We follow the settings in \citet{jieyu2017men} to focus on the imSitu vSRL dataset~\cite{yatskar2016situation}, in which we are supposed to predict the activities and roles in given images and this can be regraded as a structure prediction task (see Fig.  \ref{fig:imsitu}). 
    
    We apply the Conditional Random Field (CRF) model for the structure prediction task. We denote $\mathbf{y}$ as a joint prediction result for all instances, and $\mathbf{y}^i$ as a prediction result for instance $i$. We use $\mathbf{y}_v$ to denote the predicted activity, and $\mathbf{y}_r$ to denote the predicted role. An activity can have multiple roles and usually one of them conveys the gender information. For an instance $i$, the CRF model predicts the scores for every activity and role, and the score for a prediction is the summation of all these scores. Formally,
    $$f_\theta(\mathbf{y}^i,i)=s_\theta(\mathbf{y}^i_v,i)+\sum\nolimits_{e\in \mathbf{y}^i_r}s_\theta(\mathbf{y}^i_v,e,i),$$
    where $s_\theta(\mathbf{y}^i_v,i)$ and $s_\theta(\mathbf{y}^i_v,e,i)$ are the scores for activity $\mathbf{y}^i_v$ of instance $i$, and the score for role $e$ of instance $i$ with activity $\mathbf{y}^i_v$, respectively. We can infer the top structure for instance $i$ by:
    $$\arg\max\nolimits_{\mathbf{y}^i\in\mathcal{Y}^i}f_\theta(\mathbf{y}^i,i),$$
    where $\mathcal{Y}^i$ refers to all the possible assignments to the instance.
    
\section{Bias Amplification Quantification and Corpus-level Constraints}
\label{sec:corpusconstr}
\citet{jieyu2017men} demonstrate bias amplification in the top prediction and present a bias mitigation technique by inference with corpus-level constraints. In the following, we extend their study to analyze the bias amplification in the posterior distribution by the CRF model and define the corresponding corpus-level constraints. 

    Formally, the probability of prediction $\mathbf{y}^i$ for instance $i$ and the joint prediction $\mathbf{y}$ defined by CRF model with parameters $\theta$ are given by
    \begin{equation}
        \label{eq:crfprob}
        \begin{aligned}
            &p_\theta(\mathbf{y}^i,i)\propto \exp(f_\theta(\mathbf{y}^i, i)),\\
            &p_\theta(\mathbf{y})=\prod\nolimits_i p_\theta(\mathbf{y}^i,i),
        \end{aligned}
    \end{equation}
    since instances are mutually independent.
    
    In this section, we will define how to quantify the bias and the bias amplification in the distribution, and introduce the corpus-level constraints towards restricting the bias in the distribution.

    We focus on the gender bias on activities in the vSRL task. To quantify the gender bias given a particular activity $v^*$, \citet{jieyu2017men} uses the percentage that $v^*$ is predicted together with male agents among all prediction with genders. This evaluation focuses on the top prediction. In the contrast, we define bias function $B(p,v^*,D)$ w.r.t distribution $p$ and activity $v^*$, evaluating the bias toward male in dataset $D$ based on the conditional probability $P(X|Y)$, where $event\ Y:$ given an instance, its activity is predicted to be $v^*$ and its role is predicted to have a gender; $event\ X:$ this instance is predicted to have gender male. Formally,
    \begin{equation}
        \label{eq:rfunc}
        \begin{aligned}
             & B(p,v^*,D) \\ 
            =& \mathbb{P}_{i\sim D, \mathbf{y}\sim p}(\mathbf{y}^i_r\in M|\mathbf{y}^i_v=v^*\wedge \mathbf{y}^i_r\in M\cup W) \\
            =& \frac{\sum_{i\in D} \sum_{\mathbf{y}^i:\mathbf{y}^i_v=v^*,\mathbf{y}^i_r\in M}p(\mathbf{y}^i, i)}{\sum_{i\in D} \sum_{\mathbf{y}^i:\mathbf{y}^i_v=v^*,\mathbf{y}^i_r\in M\cup W}p(\mathbf{y}^i, i)}. \\
        \end{aligned}
    \end{equation}
    
    This bias can come from the training set $D_{tr}$. Here we use $b^*(v^*, male)$ to denote the ``dataset bias'' toward male in the training set, measured by the ratio of between male and female from the labels:
    
    $$b^*=\frac{\sum_{i\in D_{tr}} \mathbf{1}[\hat{\mathbf{y}}^i_v=v^*,\hat{\mathbf{y}}^i_r\in M]}{\sum_{i\in D_{tr}} \mathbf{1}[\hat{\mathbf{y}}^i_v=v^*,\hat{\mathbf{y}}^i_r\in M\cup W]},$$
    where $\hat{\mathbf{y}}^i$ denotes the label of instance $i$.
    
    Ideally, the bias in the distribution given by CRF model should be consistent with the bias in the training set, since CRF model is trained by maximum likelihood. However, the amplification exists in practice. Here we use the difference between the bias in the posterior distribution and in training set to quantify the bias amplification, and average it over all activities to quantify the amplification in the whole dataset:
    \begin{equation*}
        \begin{aligned}
            A(p,v^*,D)&=sgn(b^*-0.5)[B(p,v^*,D)-b^*],\\
            \bar{A}(p,D)&=\frac{1}{|V|}\sum_{v^*\in V} A(p,v^*,D).\\
        \end{aligned}
    \end{equation*}
    Note that if we use the top prediction indicator function to replace $p$ in $A,\bar{A}$, it is the same as the definition of the bias amplification in top prediction in \citet{jieyu2017men}.
    
    The corpus-level constraints aim at mitigating the bias amplification in test set $D_{ts}$ within a pre-defined margin $\gamma$, 
    \begin{equation}
        \label{eq:constr}
        \forall v^*,\ |A(p,v^*,D_{ts})| \leq \gamma.
    \end{equation}
        
\section{Posterior Regularization}
    Posterior regularization \citep{ganchev2010posterior} is an algorithm leveraging corpus-level constraints to regularize the posterior distribution for a structure model. Specifically, given corpus-level constraints and a distribution predicted by a model, we 1) define a feasible set of the distributions with respect to the constraints; 2) find the closest distribution in the feasible set from given distribution; 3) do maximum a posteriori (MAP) inference on the optimal feasible distribution.
    
    The feasible distribution set $Q$ is defined by the corpus-level constraints defined in Eq. \eqref{eq:constr}:
    \begin{equation}
        \label{eq:constrQ}
        Q=\{q\ |\  \forall v^*,\ |B(q, v^*, D_{ts}) - b^*
        |\leq \gamma\},
    \end{equation}
    where $B(\cdot)$ is defined in Eq. \eqref{eq:rfunc}. 
    
    Given the feasible set $Q$ and the model distribution $p_\theta$ defined by Eq. \eqref{eq:crfprob}, we want to find the closest feasible distribution $q^*:$
    \begin{equation}
        \label{eq:probj}
        q^*=\arg\min\nolimits_{q\in Q} KL(q\|p_\theta).
    \end{equation}
    
    This is an optimization problem and our variable is the joint distribution $q$ with constraints, which is intractable in general. Luckily, according to the results in \citet{ganchev2010posterior}, if the feasible set $Q$ is defined in terms of constraints feature functions $\phi$ and their expectations:
    \begin{equation}
        \label{eq:expQ}
        Q=\{q\ |\ \mathbb{E}_{\mathbf{y}\sim q}[\phi(\mathbf{y})\leq\mathbf{c}]\},
    \end{equation}
    Eq. \eqref{eq:probj} will have a close form solution 
    \begin{equation}
        \label{eq:qstar}
        q^*(\mathbf{y})=\frac{p_\theta(\mathbf{y})\exp(-\lambda^*\cdot \phi(\mathbf{y}))}{Z(\lambda^*)},
    \end{equation}
    where $\lambda^*$ is the solution of
    \begin{equation}
        \label{eq:lambda}
        \begin{aligned}
            \lambda^* &= \arg\max\nolimits_{\lambda\geq 0}-\mathbf{c}\cdot \lambda-\log Z(\lambda). \\
            Z(\lambda) &= \sum\nolimits_{\mathbf{y}} p_\theta(\mathbf{y})\exp(-\lambda\cdot \phi(\mathbf{y})).
        \end{aligned}
    \end{equation}

    Actually, we can derive the constraints into the form we want. We set $\mathbf{c}=\mathbf{0}$ and
    \begin{equation}
         \label{eq:phi}
         \phi(\mathbf{y})=\sum\nolimits_i \phi^i(\mathbf{y}^i).
    \end{equation}
    We can choose a proper $\phi^i(\mathbf{y}^i)$ to make Eq. \eqref{eq:constrQ} equal to Eq. \eqref{eq:expQ}. The detailed derivation and the definition of $\phi^i(\mathbf{y}^i)$ are shown in Appendix \ref{app:phi}.
    
    We can solve Eq. \eqref{eq:lambda} by gradient-based methods to get $\lambda^*$, and further compute the close form solution in Eq. \eqref{eq:qstar}. Actually, considering the relation between $\mathbf{y}$ and $\mathbf{y}^i$ in Eq. \eqref{eq:crfprob} and \eqref{eq:phi}, we can factorize the solution in Eq. \eqref{eq:qstar} on instance level:
    \begin{equation*}
        q^*(\mathbf{y}^i,i)=\frac{p_\theta(\mathbf{y}^i,i)\exp(-\lambda^*\cdot \phi^i(\mathbf{y}^i))}{Z^i(\lambda^*)},
    \end{equation*}
    and the derivation details are in Appendix \ref{app:expectation}. With this, we can reuse original inference algorithm to conduct MAP inference based on the distribution $q^*$ for every instance seperately.
           
    \begin{figure*}[t]
        \centering
        \begin{subfigure}{0.45\linewidth}
            \includegraphics[width=\linewidth]{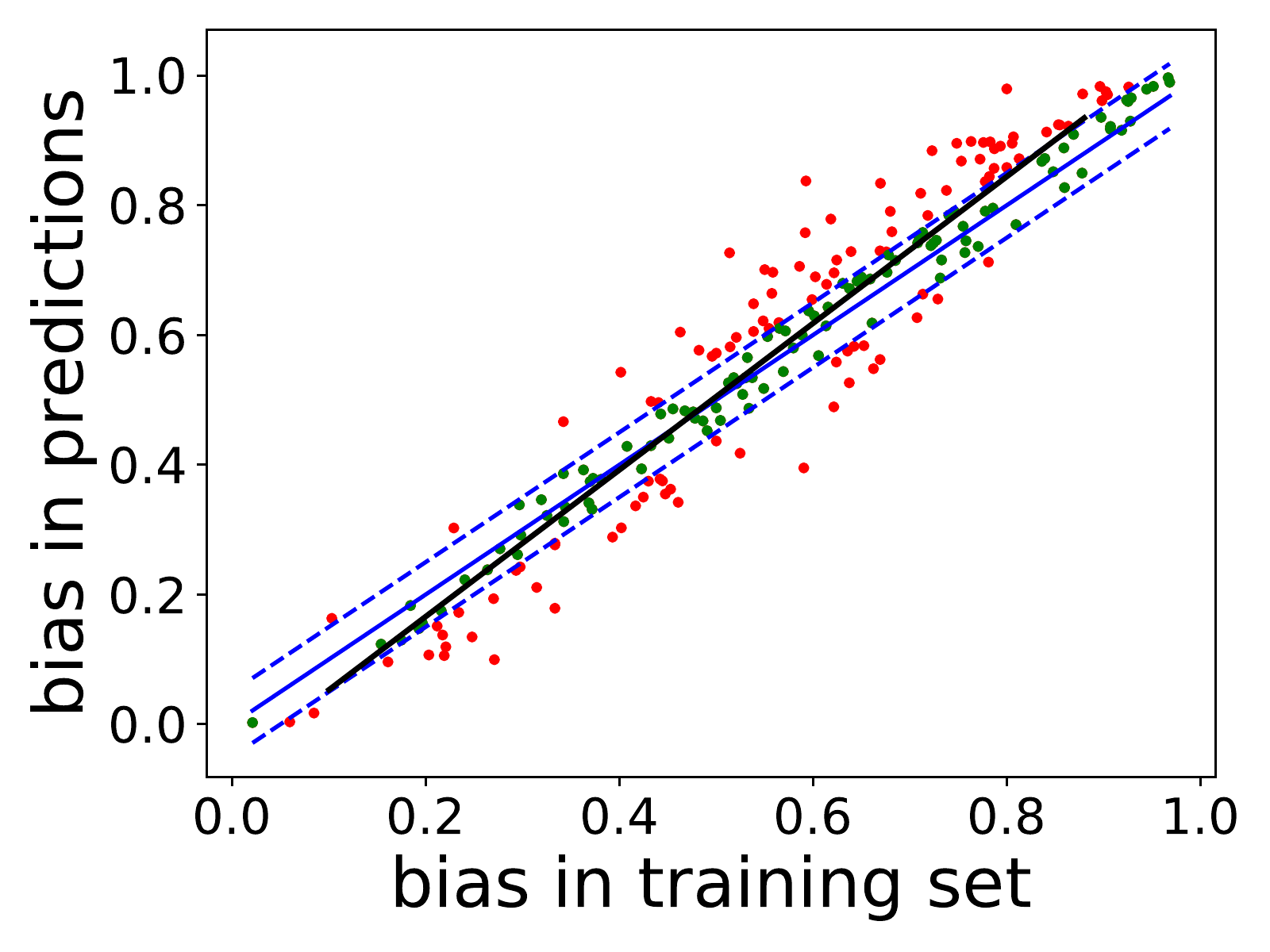}
            \subcaption{bias in distribution before bias mitigation.}
            \label{fig:dist_before}
        \end{subfigure}
        \hspace{0.2in}
        \begin{subfigure}{0.45\linewidth}
            \includegraphics[width=\linewidth]{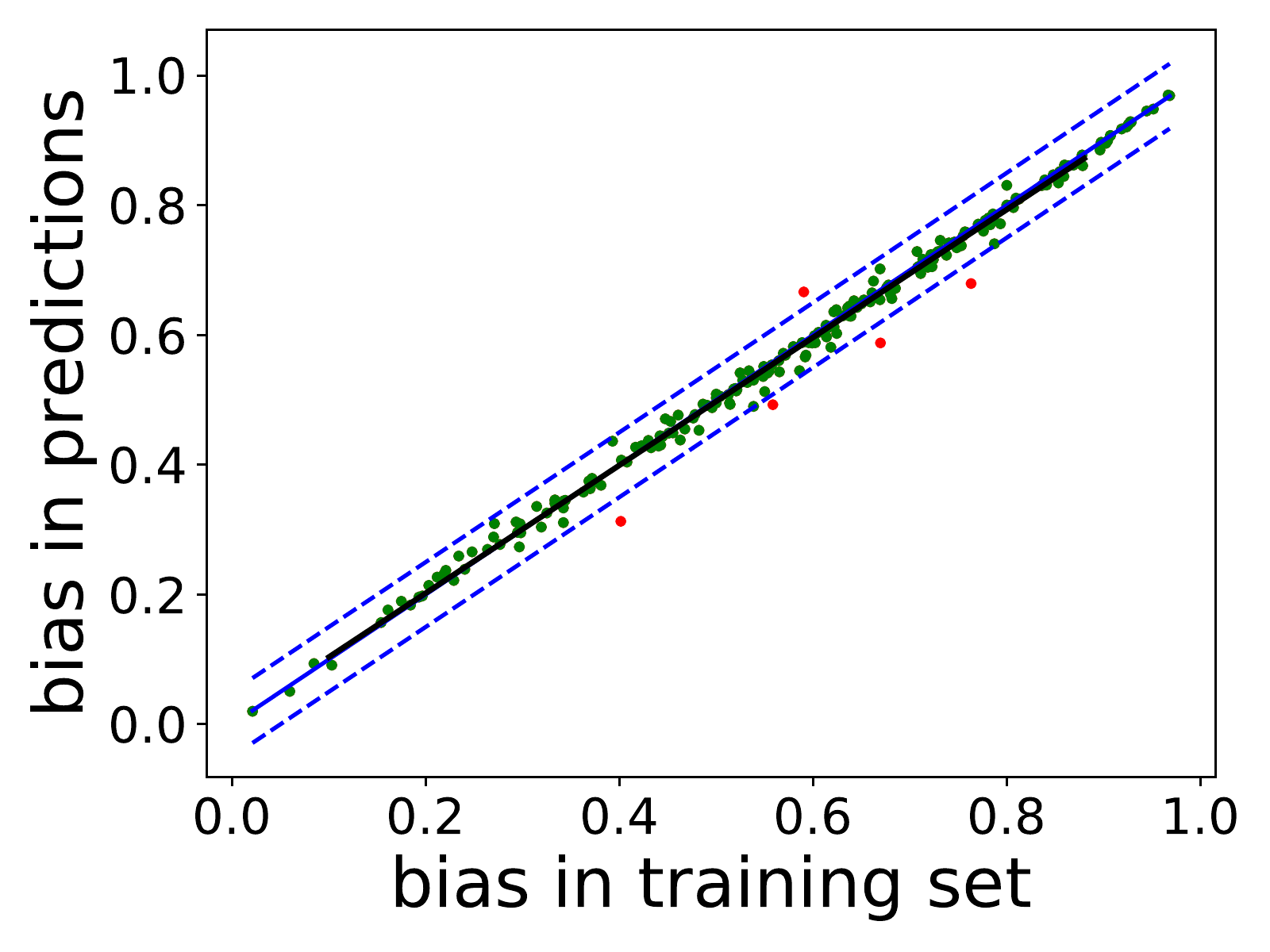}
            \subcaption{bias in distribution after bias mitigation.}
            \label{fig:dist_after}
        \end{subfigure}
        \\
        \centering
        \begin{subfigure}{0.45\linewidth}
            \includegraphics[width=\linewidth]{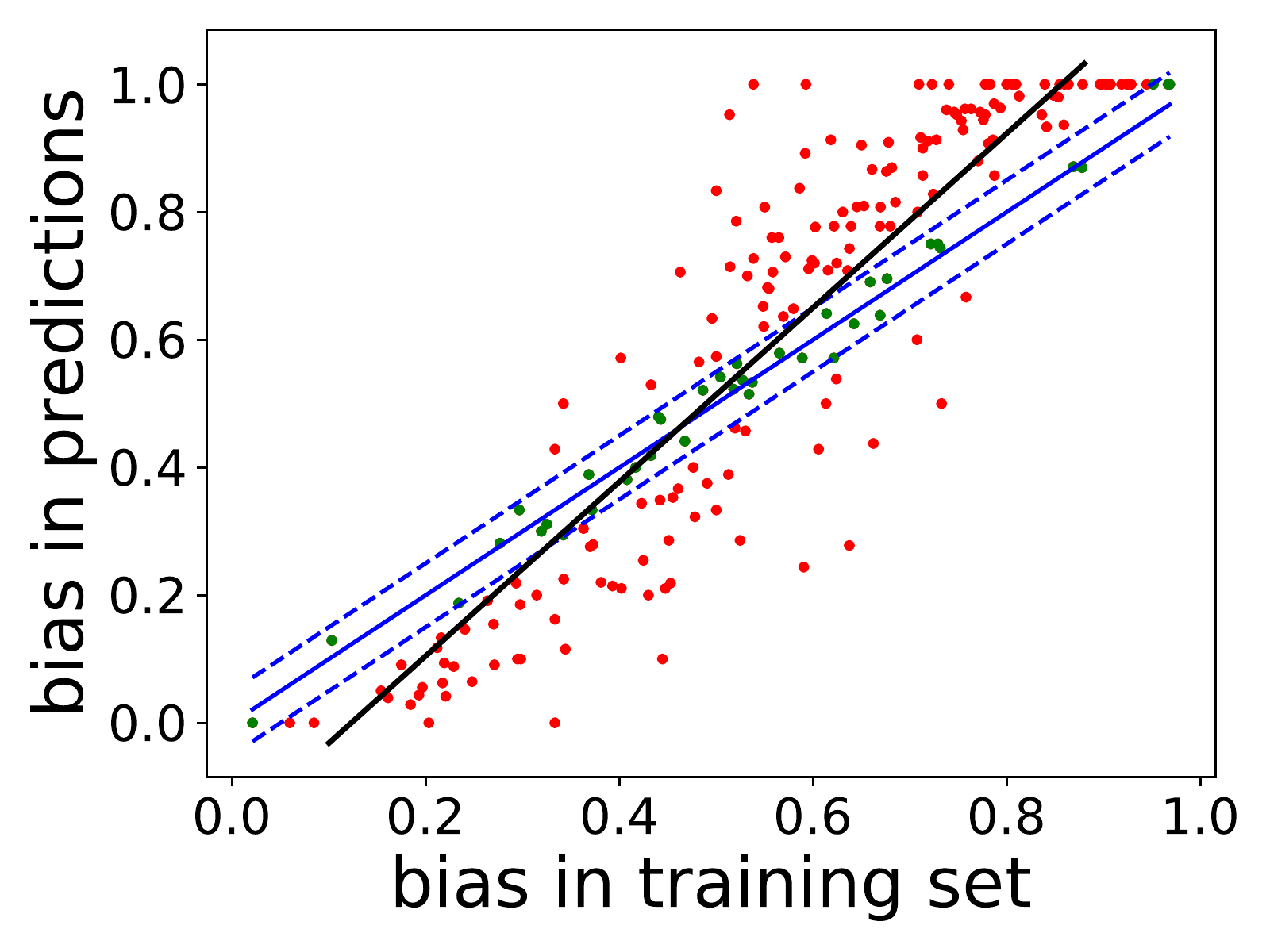}
            \subcaption{bias in top predictions before bias mitigation.}
            \label{fig:inf_before}
        \end{subfigure}
        \hspace{0.2in}
        \begin{subfigure}{0.45\linewidth}
            \includegraphics[width=\linewidth]{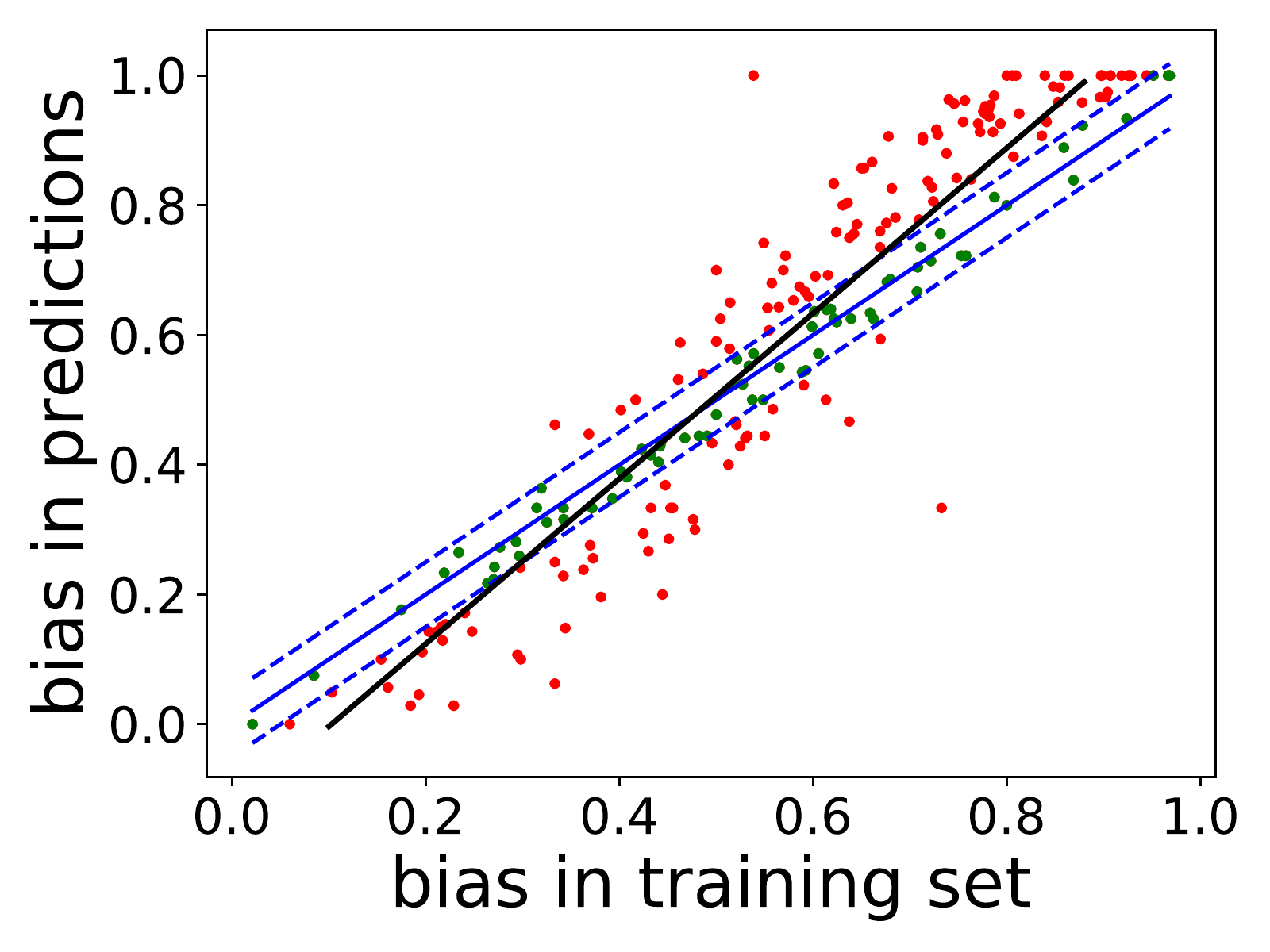}
            \subcaption{bias in top predictions after bias mitigation.}
            \label{fig:inf_after}
        \end{subfigure}
        \caption{x-axis and y-axis are the bias toward male in the training corpus and the predictions, respectively.
        Each dot stands for an activity. The blue reference lines indicate the bias score in training is equal to that in test and the dash lines indicate the margin $(=0.05)$. The dots in red stand for being out of margin and violating the constraints.
        The black lines are linear regressions of the dots. Results show that we can almost remove the bias amplification in distributions (see \ref{fig:dist_before} and \ref{fig:dist_after}), and reduce 30.9\% amplification in top predictions (see \ref{fig:inf_before} and \ref{fig:inf_after}) after applying posterior regularization.}
        \label{fig:before}
    \end{figure*}
        
\section{Experiments}
    We conduct experiments on the vSRL task to analyze the bias amplification issue in the posterior distribution and demonstrate the effectiveness of the proposed bias mitigation technique. 
    \paragraph{Dataset} 
        Our experiment settings follow \citet{jieyu2017men}. We evaluate on imSitu \citep{yatskar2016situation} that activities are selected from verbs, roles are from FrameNet \citep{baker1998framenet} and nouns from WordNet \citep{fellbaum1998wordnet}. We filter out the non-human oriented verbs and images with labels that do not indicate the genders. 
    \paragraph{Model} 
        We analyze the model purposed together with the dataset. The score functions we describe in Sec. \ref{sec:background} are modeled by VGG \citep{simonyan2015vgg} with a feedforward layer on the top of it. The scores are fed to CRF for inference.
            
    \subsection{Bias Amplification in Distribution}
        Figures \ref{fig:dist_before} and \ref{fig:inf_before} demonstrate the bias amplification in both posterior distribution $p_\theta$ and the top predictions $\mathbf{y}$ defined in Sec.\ref{sec:corpusconstr}, respectively. For most activities with the bias toward male (i.e., higher bias score) in the training set, both the top prediction and posterior distribution are even more biased toward male, vise versa. If the bias is not amplified, the dots should be scattered around the reference line. However, most dots are on the top-right or bottom-left, showing the bias is amplified. The black regression line with $slope>1$ also indicates the amplification. Quantitatively, $109$ and $173$ constraints are violated when analyzing the bias in distribution an in top predictions. 
        
        Most recent models are trained by minimizing the cross-entropy loss which aims at fitting the model's predicted distribution with observed distribution on the training data. In the inference time, the model outputs the top predictions based on the underlying prediction distribution. Besides, in practice, the distribution has been used as an indicator of confidence in the prediction. Therefore, understanding bias amplification in distribution provides a better view about this issue.
        
        \begin{figure}[t]
            \centering
            \includegraphics[width=\linewidth]{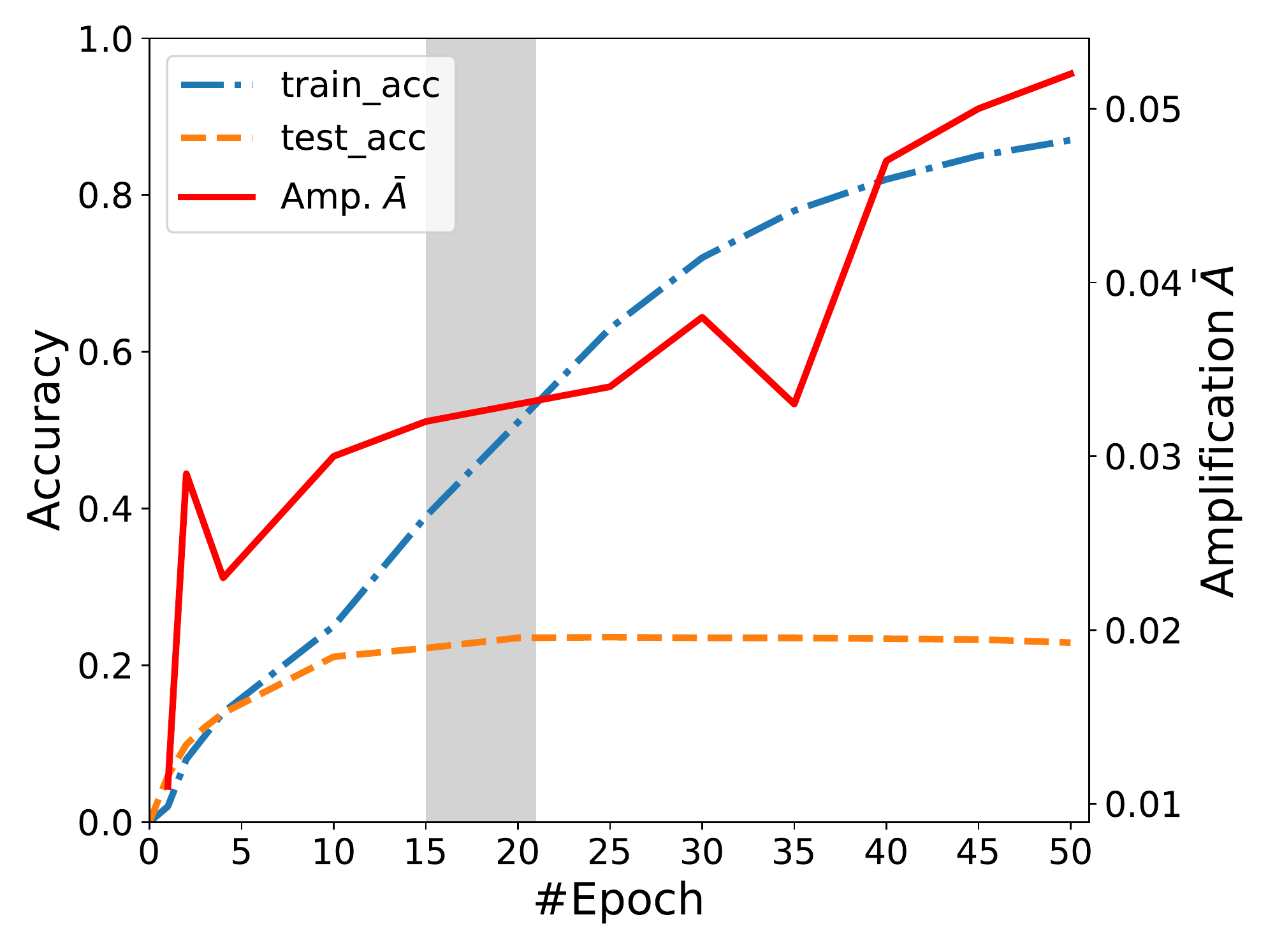}
            \caption{The curve of training and test accuracy, and bias amplification with the number of training epochs. The optimal model evaluated on the development set is found in the grey shade area.}
            \label{fig:early}
        \end{figure}
    
        To analyze the cause of bias amplification, we further show the degree of amplification along with the learning curve of the model (see Fig. \ref{fig:early}). We observed that when the model is overfitted, the distribution of the model prediction becomes more peaky\footnote{This effect, called overconfident, has been also discussed in the literature~\citep{guo2017calibration}.}. 
        We suspect this is one of the key reasons causes the bias amplification. 
       
        
        \subsection{Bias Amplification Mitigation} 
            We set the margin $\gamma=0.05$ for every constraint in evaluation. However, we employ a stricter margin ($\gamma=0.001$) in performing posterior regularization to encourage the model to achieve a better feasible solution. 
            We use mini-batch to estimate the gradient w.r.t $\lambda$  with Adam optimizer \citep{kingma2014adam} when solving Eq. \eqref{eq:probj}. We set the batchsize to be $39$ and train for $10$ epochs. The learning rate is initialized as $0.1$ and decays after every mini-batch with the decay factor $0.998.$
        
    \paragraph{Results}

        We then apply the posterior regularization technique to mitigate the bias amplification in distribution. Results are demonstrated in Figures \ref{fig:dist_after} (distribution) and \ref{fig:inf_after} (top predictions). The posterior regularization effectively calibrates the bias in distribution and only $5$ constraints are violated after the calibration. 
        The average bias amplification is close to $0$ ($\bar{A}$: $0.032$ to $-0.005$). By reducing the amplification of bias in distribution, the bias amplification in top predictions also reduced by 30.9\% ($\bar{A}$: $0.097$ to $0.067$). At the same time, the model's performance is kept (accuracy: $23.2\%$ to $23.1\%$). 
        
        Note that calibrating the bias in distribution cannot remove all bias amplification in the top predictions. 
        We posit that the requirement of making hard predictions (i.e., maximum a posteriori estimation) also amplifies the bias when evaluating the top predictions.  

\section{Conclusion}
    We analyzed the bias amplification from the posterior distribution perspective, which provides a better view to understanding the bias amplification issue in natural language models as these models are trained with the maximum likelihood objective. We further proposed a bias mitigation technique based on posterior regularization and show that it effectively reduces the bias amplification in the distribution.     Due to the limitation of the data, we only analyze the bias over binary gender. However, our analysis and the mitigation framework is general and can be adopted to other applications and other types of bias.
    
    One remaining open question is why the gender bias in the posterior distribution is amplified. 
    We posit that the regularization and the over-fitting nature of deep learning models might contribute to the bias amplification. However, a comprehensive study is required to prove the conjecture and we leave this as future work.

    
    

\paragraph{Acknowledgement} This work was supported in part by National Science Foundation Grant IIS-1927554. We thank anonymous reviewers and members of the UCLA-NLP lab for their feedback.

\bibliography{acl2020}
\bibliographystyle{acl_natbib}

\input{appendix.tex}

\end{document}

%% file: appendix.tex







\newpage
\appendix

    
\section{Definition of the Feature Functions}
\label{app:phi}
    The feature function for predictions $\mathbf{y}$ is defined as the summation of feature functions for each instance $\mathbf{y}^{i}$, which is a $2n-$dimensional vector where $n$ is the number of constraints. Each entry is the feature function corresponding to a constraint and the inequality sign direction. Formally,
    \begin{equation*}
        \begin{aligned}
            \phi^i_{v^*,-}(\mathbf{y}^i)\!&=\!
            \begin{cases}
                1-b^*-\gamma & \mathbf{y}^i_v=v^*,\mathbf{y}^i_r\in M\\
                -b^*-\gamma & \mathbf{y}^i_v=v^*,\mathbf{y}^i_r\in W\\
                0 & otherwise\\
            \end{cases}\\
            \phi^i_{v^*,+}(\mathbf{y}^i)\!&=\!
            \begin{cases}
                -1+b^*-\gamma & \mathbf{y}^i_v=v^*,\mathbf{y}^i_r\in M\\
                b^*-\gamma & \mathbf{y}^i_v=v^*,\mathbf{y}^i_r\in W\\
                0 & otherwise\\
            \end{cases}\\
            \phi^i&=(\phi^i_{v_1,-},\phi^i_{v_1,+},...,\phi^i_{v_n,-},\phi^i_{v_n,+})\\
            \phi(\mathbf{y})&=\sum_i \phi^i(\mathbf{y}^i)\\
        \end{aligned}
    \end{equation*}
\section{Derivation of Feature Functions Expectation}
\label{app:expectation}
    We can derive the feature functions expection as
    \begin{eqnarray*}
        \mathbb{E}_{\mathbf{y}\sim q}[\phi(\mathbf{y})] &\leq& \mathbf{0} \\
        \mathbb{E}_{\mathbf{y}\sim q}\left[\sum_i \phi^i(\mathbf{y}^i)\right] &\leq& \mathbf{0} \\
        \sum_i \mathbb{E}_{\mathbf{y}^i\sim q(\cdot,i)}\left[\phi^i(\mathbf{y}^i)\right] &\leq& \mathbf{0} \\
    \end{eqnarray*}
    Thus, it is equivalent as $\forall v^*,$ 
    $$\sum_i \mathbb{E}_{\mathbf{y}^i\sim q(\cdot,i)}\left[\phi^i_{v^*,-}(\mathbf{y}^i)\right] \leq \mathbf{0},$$
    $$\sum_i \mathbb{E}_{\mathbf{y}^i\sim q(\cdot,i)}\left[\phi^i_{v^*,+}(\mathbf{y}^i)\right] \leq \mathbf{0}.$$
    The inequality about $\phi^i_{v^*,-}$ can be derived as
    \begin{eqnarray*}
        \sum_i \mathbb{E}_{\mathbf{y}^i\sim q(\cdot,i)}\left[\phi^i_{v^*,-}(\mathbf{y}^i)\right] &\leq& \mathbf{0} \\
        \sum_i \sum_{\mathbf{y}^i} q(\mathbf{y}^i,i)\phi^i_{v^*,-}(\mathbf{y}^i) &\leq& \mathbf{0} \\
        \sum_i\sum_{\mathbf{y}^i:\mathbf{y}^i_v=v^*,\mathbf{y}^i_r\in M} (1-b^*-\gamma)q(\mathbf{y}^i,i) &-& \\
        \sum_i\sum_{\mathbf{y}^i:\mathbf{y}^i_v=v^*,\mathbf{y}^i_r\in W} (b^*+\gamma)q(\mathbf{y}^i,i)&\leq& \mathbf{0} \\
        \frac{\sum_i \sum_{\mathbf{y}^i:\mathbf{y}^i_v=v^*,\mathbf{y}^i_r\in M}q(\mathbf{y}^i, i)}{\sum_i \sum_{\mathbf{y}^i:\mathbf{y}^i_v=v^*,\mathbf{y}^i_r\in M\cup W}q(\mathbf{y}^i, i)} &\leq& b^*+\gamma \\
        B(q,v^*,\cdot) \leq b^*+\gamma
    \end{eqnarray*}
    The inequality about $\phi^i_{v^*,-}$ can be derived similarly.
    
